\pdfoutput=1
\documentclass[11pt]{article}
\usepackage{adjustbox}
\usepackage{amsmath} % or \usepackage{bm}
\usepackage{booktabs}
\usepackage{amssymb}
\usepackage{emnlp2023}

\usepackage{times}
\usepackage{latexsym}

\usepackage[T1]{fontenc}
\usepackage[utf8]{inputenc}

\usepackage{microtype}

 \usepackage{multirow} 
\title{SurreyAI 2023 Submission for the Quality Estimation Shared Task}

\author{Archchana Sindhujan\textsuperscript{*1},  Diptesh Kanojia\textsuperscript{1},  Constantin Orasan\textsuperscript{2} and  Tharindu Ranasinghe\textsuperscript{3} \\
\textsuperscript{1} Institute for People-Centred AI, University Of Surrey , UK \\  \textsuperscript{2}Centre for Translation Studies, University of Surrey, UK \\ \textsuperscript{3}
School of Computer Science and Digital Technologies, Aston University, UK \\
\texttt{\{a.sindhujan,d.kanojia,c.orasan\}@surrey.ac.uk,} \\
\texttt{ t.ranasinghe@aston.ac.uk}
}

\begin{document}
\maketitle
\begin{abstract}
Quality Estimation (QE) systems are important in situations where it is necessary to assess the quality of translations, but there is no reference available. This paper describes the approach adopted by the SurreyAI team for addressing the Sentence-Level Direct Assessment shared task in WMT23. The proposed approach builds upon the TransQuest framework, exploring various autoencoder pre-trained language models within the MonoTransQuest architecture using single and ensemble settings. The autoencoder pre-trained language models employed in the proposed systems are XLMV, InfoXLM-large, and XLMR-large. The evaluation utilizes Spearman and Pearson correlation coefficients, assessing the relationship between machine-predicted quality scores and human judgments for 5 language pairs (English-Gujarati, English-Hindi, English-Marathi, English-Tamil and English-Telugu). The MonoTQ-InfoXLM-large approach emerges as a robust strategy, surpassing all other individual models proposed in this study by significantly improving over the baseline for the majority of the language pairs.

\end{abstract}

\section{Introduction}

The primary objective of quality estimation (QE) systems is to assess the quality of a translation without relying on a reference translation. This make QE valuable within translation processes, as it enables the determination of whether an automatically generated translation is sufficiently accurate for a specific purpose. This aids in deciding whether the translation can be used as is, requires human intervention for full translation, or necessitates post-editing by a human translator ~\citep{kepler-etal-2019-openkiwi}. Quality estimation can be conducted across various levels: word/phrase level, sentence level and document level. This paper considers only the sentence-level QE and presents our participation in the WMT23 Sentence-level direct assessment (DA) shared task. 
% I suggest the following text instead
In the context of this task, participating systems are required to predict the DA score for a given (source, target) pair. This score serves as a measure of the translation quality.

Building upon the ideas presented in TransQuest by ~\citet{ranasinghe2020transquest}, our investigation explores the use of various pre-trained models within the MonoTransQuest architecture for the sentence-level quality estimation shared task. The architecture employs autoencoder pre-trained language models to fine-tune the QE data to predict a score which indicates the quality of translation. Using the MonoTransQuest architecture as the base we employ the pre-trained transformers separately to implement the systems MonoTQ-XLMV, MonoTQ-InfoXLM-large and MonoTQ-XLMR-large. In addition, we propose ensembleTQ which combines the output of MonoTransQuest when using different pre-trained models.  All the proposed systems achieve a significantly higher Spearman correlation score compared to the baseline.

The paper is structured as follows. Section 2 briefly presents related work on quality estimation. Section 3 provides a concise overview of the dataset used in the sentence-level QE shared task. Moving on to Section 4, we introduce the autoencoder pre-trained language models and proposed systems and detail the training methodology. Section 5 is dedicated to the evaluation and Section 6 comprises the result and discussion. The paper concludes by summarizing findings, highlighting conclusions, and suggesting potential avenues for future research in the final section.

\section{Related work}
Quality estimation in machine translation has evolved significantly throughout the years. Initially, it relied on feature engineering and conventional machine learning techniques like SVM and basic neural networks~\citep{specia-etal-2015-multi, scarton-specia-2014-document}. However, Neural Networks has since become central to quality estimation, where there is no more need of feature engineering, and the models can be trained directly on the data ~\citep{kepler-etal-2019-openkiwi, kepler-etal-2019-unbabels, specia-etal-2018-findings}. Recently, Transformer-based architectures have arisen as robust solutions for machine translation and quality estimation. Notably, there are two widely recognized frameworks that leverage this transformative approach for QE tasks: TransQuest  ~\cite {ranasinghe-etal-2020-transquest-wmt2020} and CometKIWI ~\cite {rei-etal-2022-cometkiwi}.

Ensemble methods have also been explored extensively in Quality Estimation tasks~\citep{ bao-etal-2022-alibaba, geng-etal-2022-njunlps, kepler-etal-2019-openkiwi, ranasinghe2020transquest, rei-etal-2022-cometkiwi}. 
The ensemble approach from \citet{lim-park-2022-papagos} using K-folds consistently outperformed the standard method, underscoring the prevalent belief that ensemble strategies enhance performance outcomes. However, some of the research studies~\cite{ranasinghe2020transquest, bao-etal-2022-alibaba,rei-etal-2022-cometkiwi} show that combining multi-lingual models through ensembling yields better results than the traditional k-fold ensemble technique. \citet{geng-etal-2022-njunlps} suggest an alternative ensemble method that merges the results from models trained using various sentence-level metrics.

Our study delves into the performance of cutting-edge pre-trained transformer-based approaches when applied to sentence-level Quality Estimation tasks.
 
\section{Dataset}

We focus on Sentence-Level Direct Assessment tasks which comprise datasets for 5 language pairs which has English on the source side and Indian languages on the target side: English-Gujarati (En-Gu), English-Hindi (En-Hi), English-Marathi (En-Mr), English-Tamil (En-Ta) and English-Telugu (En-Te). Among these language pairs, En-Hi language pair is considered mid-resourced and all the other language pairs are low-resourced. Each language pair includes around 7,000 sentence pairs in the training set, as well as around 1,000 sentence pairs in both the development and testing sets. Each translation was evaluated by three professional translators who assigned a score between 0 and 100. These Direct Assessment (DA) scores were normalized using the z-score. The final score for the sentence-level task requires predicting the mean DA z-scores for the test sentence pairs. More details on this can be found in \citet{zerva-etal-2022-findings}.

\section{Methodology}

This section outlines the approach taken to formulate our quality estimation techniques. We begin by detailing the autoencoder pre-trained language models employed in our architecture. Then we explain the architecture and the strategy employed to train these network architectures in detail.

 \begin{figure*}[t]
 \centering
  \includegraphics[width=0.85\textwidth]{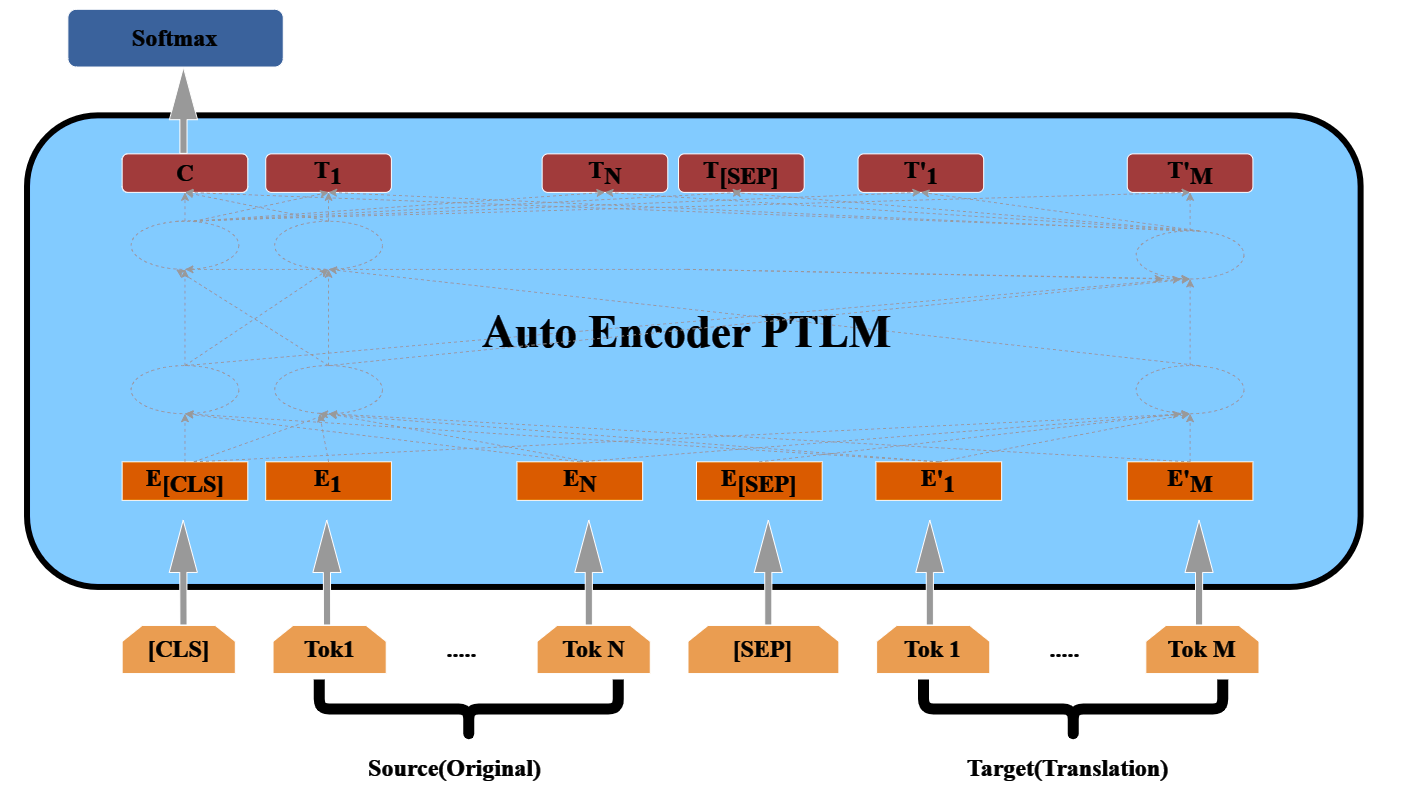}
\caption{ Architecture diagram of the proposed approaches}
\label{fig2}       % Give a unique label
\end{figure*}

\subsection{Pre-trained models for fine-tuning}

\begin{enumerate}

\item  XLMR-large \\ \\
XLM-Roberta \citep{conneau-etal-2020-unsupervised} is a pre-trained transformer-based language model which is a part of the Cross-lingual Language Model (XLM). This model employs large-scale cross-lingual pre-training to capture contextual information and representations across 100 languages.  The model is trained on 2.5TB of filtered CommonCrawl data from multiple languages, allowing it to effectively learn cross-lingual and language-specific patterns. The XLM-R architecture takes sequences as input, with a maximum token limit of 512, and generates contextualized embeddings for each token, enabling it to perform well on various natural language processing tasks across different languages \cite{ranasinghe2020transquest, ranasinghe-etal-2021-exploratory}.

\item  XLMV \\ \\
XLMV is a multilingual language model with a one million token vocabulary trained on 2.5TB of data from Common Crawl (same as XLM-R) \citep{Liang2023XLMVOT}. In the context of large multilingual language models, a common practice involves employing a single vocabulary shared across a diverse set of languages. Even with the expansion in model complexity, including parameter count and depth, the vocabulary size has remained relatively static. This constraint in vocabulary hampers the potential of multilingual models such as XLM-R to capture nuanced representations effectively ~\citep{wang-etal-2019-improving}. 

XLMV introduced an innovative strategy addressing this issue by achieving scalability to extensive multilingual vocabularies. XLMV involves prioritizing vocabulary allocation based on language-specific lexical overlap, ensuring sufficient coverage for each language. The outcome is tokenizations that hold enhanced semantic significance and are generally more concise compared to those generated by XLM-R. 

\item InfoXLM-large \\ \\
InfoXLM-large ~\citep{chi-etal-2021-infoxlm}, is an information-theoretic framework for cross-lingual language model pre-training. It extends the XLM-R architecture by formulating cross-lingual pre-training to maximize mutual information between multilingual texts at different granularities. This approach enhances the model's capability to learn effective cross-lingual representations by capturing shared information across languages. InfoXLM-large introduces a novel pre-training task based on contrastive learning, treating bilingual sentence pairs as views of the same meaning. By jointly training on monolingual and parallel corpora, the model improves the transferability of its representations for various downstream cross-lingual tasks \cite{rei-etal-2022-cometkiwi,bao-etal-2022-alibaba}.

\end{enumerate}

\begin{table*}
\centering
\begin{adjustbox}{width=1\textwidth}
\small
\begin{tabular}{clcccccccccc}
\toprule
 &  & \multicolumn{2}{c}{\textbf{En-Gu}} &  \multicolumn{2}{c}{\textbf{En-Hi}} & \multicolumn{2}{c}{\textbf{En-Mr}} &
\multicolumn{2}{c}{\textbf{En-Ta}} &
\multicolumn{2}{c}{\textbf{En-Te}} \\
\cmidrule(lr){3-4} \cmidrule(lr){5-6} \cmidrule(lr){7-8} \cmidrule(lr){9-10} \cmidrule(lr){11-12}
 &  \textbf {Method}  & $\rho$ & r  & $\rho$ & r & $\rho$ & r & $\rho$ & r & $\rho$ & r\\
\midrule

I & Baseline  & 0.337 & 0.307 & 0.281 & 0.245  & 0.392 & 0.427  & 0.507 & 0.402  & 0.193 & 0.153 \\

\\
II & MonoTQ-XLMV & 0.673 & 0.536 & 0.572 & 0.687 & \textbf{0.642} & 0.425 & 0.670 & 0.559 & \textbf{0.464} & 0.642 \\

\\
III & MonoTQ-InfoXLM-large & \textbf{0.713} & 0.656 & \textbf{0.624} & 0.726 & 0.470 & 0.030 & \textbf{0.726} & 0.662 & 0.462 & 0.719\\
\\

IV & MonoTQ-XLMR-large & 0.438 & 0.299 & 0.440 & 0.430 & 0.395 & -0.117 & 0.482 & 0.454 & 0.345 & 0.211\\

\\
V & ensembleTQ & 0.649 & 0.700 & 0.551 & 0.668 & 0.596 & 0.668 & 0.674 &  0.710 & 0.349 & 0.376 \\
\\
\bottomrule
\end{tabular}
\end{adjustbox}

\caption{\label{table_mymodels}
Spearman ($\rho$) and Pearson (r) correlation between the proposed approach predictions and human DA judgments. The best Spearman score obtained
for each language pair (any method) is marked in bold. Rows II, III, and IV indicate the single-configuration settings of
MonoTransQuest architecture with different pre-trained transformer models as explained in Section 5.1, and ensembleTQ in row V is explained in Section 5.2. The baseline results are in Row I.}
\end{table*}
\subsection{Architecture}

The proposed architecture of MonoTransQuest employs a pre-trained language model as shown in Figure 1. The MonoTransQuest architecture in TransQuest ~\citep{ranasinghe2020transquest} considers only the XLMR transformer model. In our proposed system, we train multiple multilingual QE models by fine-tuning autoencoder pre-trained language models (PTLMs) and report mean z-scores. The PTLMs are namely  XLMV, InfoXLM-large, and XLMR-large which we have explained in section 4.1.  The model's input consists of the original sentence (source) and its translation (target) concatenated, with a [SEP] token. This token marks the separation of the original sentence and the translated sentence. The pre-trained auto-encoder accepts input sequences with a token limit of 512 and produces a sequence representation as output. The initial token of the sequence is [CLS] token, encompassing a distinctive embedding to signify the entire sequence. Subsequently, embeddings are assigned to each word in the sequence. \citet{ranasinghe2020transquest} highlights the superiority of the CLS-strategy over the MEAN-strategy (calculating the mean of all output vectors corresponding to the input words) and MAX-strategy (determining the maximum value across the output vectors of input words) for pooling within the MonoTransQuest framework. We have used the CLS-strategy (using the output of the [CLS] token)  to extract the output from the transformer model. Consequently, we employed the [CLS] token's embedding as input for a softmax layer. The softmax layer predicts the translation's quality score. The mean-squared-error loss function was used as the objective function for training.

\subsection{Training and Implementation Details}
We started the training with MonoTQ-XLMV which incorporates the XLMV-base model with MonoTransQuest for all 5 language pairs. We had the batch size as 8. We have used Adam Optimizer ~\citep{kingma2014adam} with a learning rate of 2e-5. The model is trained using 3 epochs. The training process exclusively utilized the training data. Early stopping was enforced if the evaluation loss failed to show improvements over ten consecutive evaluation rounds. We continued the training with the same set of configurations for MonoTQ-InfoXLM-large and MonoTQ-XLMR-large separately. MonoTQ-XLMV and MonoTQ-InfoXLM-large required twice the training time compared to MonoTQ-XLMR-large which required approximately 40 minutes of training on a GPU with 48GB of memory.

The proposed systems are built upon the most up-to-date version of \verb|TransQuest|\footnote{https://github.
com/tharindudr/transQuest} framework and executed using Python 3.9 and PyTorch 2.0.1. The integration of pre-trained encoders (\verb|XLMV|\footnote{https://huggingface.co/facebook/xlm-v-base}, \verb|XLMR-large|\footnote{https://huggingface.co/xlm-roberta-large} and \verb|InfoXLM-large|\footnote{https://huggingface.co/microsoft/infoxlm-large}) into the MonoTransQuest architecture was facilitated through the application of HuggingFace's Transformers library.
% Footnotes are inserted with the \verb|\footnote| command.

\section{Evaluation}
In this section, we outline the evaluation outcomes of our models. We assess the performance of the proposed models under two circumstances: single model configuration and ensembleTQ. 

The primary evaluation criterion employed was Spearman's rank correlation coefficient~\citep{sedgwick2014spearman}, which is a statistical measure used to evaluate the strength and direction of association between two variables. Also, we have calculated the Pearson correlation coefficient~\citep{cohen2009pearson} as a secondary metric for the evaluation. In the context of Quality Estimation (QE) for machine translation, it is used to evaluate the correlation between the machine-predicted quality scores and the gold standard labels provided by human annotators in the test dataset. Spearman's rank correlation coefficient assesses the monotonic relationship between the two variables, unlike the Pearson correlation~\citep{cohen2009pearson}, which measures the linear relationship between two variables. It is calculated by first ranking the values of both variables in ascending or descending order and then computing the Pearson correlation coefficient between the two sets of ranks. Spearman's rank correlation coefficient is often preferred because it is less sensitive to outliers and non-linear relationships between the predicted scores and human scores. 

\begin{table*}
\centering
\begin{adjustbox}{width=1\textwidth}
\small
\begin{tabular}{clcccccccccc}
\toprule
 &  & \multicolumn{2}{c}{\textbf{En-Gu}} &  \multicolumn{2}{c}{\textbf{En-Hi}} & \multicolumn{2}{c}{\textbf{En-Mr}} &
\multicolumn{2}{c}{\textbf{En-Ta}} &
\multicolumn{2}{c}{\textbf{En-Te}} \\
\cmidrule(lr){3-4} \cmidrule(lr){5-6} \cmidrule(lr){7-8} \cmidrule(lr){9-10} \cmidrule(lr){11-12}
 &  \textbf {Team}  & $\rho$ & r & $\rho$ & r & $\rho$ & r & $\rho$ & r & $\rho$ & r\\
\midrule

1 & Unbabel-IST  & \textbf{0.714} & \textbf{0.745} & 0.598 & 0.667  & \textbf{0.704 }& \textbf{0.735}  & 0.739 & 0.733  & 0.388 & 0.362 \\

\\
2 & IOL Research & 0.695 & 0.742 & 0.6 & 0.667 & 0.505 &  0.372 & 0.74 & 0.742 & 0.376 & 0.344 \\

\\
3 & HW-TSC & 0.691 & 0.714 & \textbf{0.644} & \textbf{0.72} & 0.692 & 0.718 & \textbf{0.775} &\textbf{ 0.778} & \textbf{0.394} & 0.35\\
\\

4 & MMT & 0.54 & 0.581 & 0.494 & 0.57 & 0.65 & 0.663 & 0.547 & 0.531 & 0.337 & 0.281\\

\\
5 & Baseline & 0.337 & 0.307 & 0.281 & 0.245 & 0.392 & 0.427 & 0.507 & 0.402 & 0.193 & 0.153\\
\\
6 & SurreyAI-ensembleTQ & 0.649 & 0.700 & 0.551 & 0.668 & 0.596 & 0.668 & 0.674 &  0.710 & 0.349 & \textbf{0.376} \\
\\
\bottomrule
\end{tabular}
\end{adjustbox}

\caption{\label{tab2_allmodels}
Spearman ($\rho$) and Pearson (r) correlation between the predictions from the participated systems in WMT23 sentence-level QE shared task and human DA judgments. The best Spearman and Pearson score obtained
for each language pair is marked in bold. Even though we have experimented with the single model configurations, we only submitted our ensembled approach (SurreyAI-ensembleTQ) for the shared task competition.}
\end{table*}

\subsection{Single model configurations}
Initially, our evaluation focused on the single model configurations of the proposed framework. This involved training a quality estimation model using a single autoencoder pre-trained language model on the training data for each language pair separately. Subsequently, we assessed each model's performance (MonoTQ-XLMV, MonoTQ-InfoXLM-large, MonoTQ-XLMR-large) on the corresponding test set for each language pair. The outcomes of this evaluation for the single model configuration are presented in Table \ref{table_mymodels}.

\subsection{EnsembleTQ}
Recently, ensemble techniques have demonstrated their efficacy in enhancing transformer-based models' performance ~\citep{xu2020improving}. Following this approach, we employed an ensemble strategy to experiment further to see whether it enhance the performance. For every input within the test set, we aggregate the output scores from various distinct pre-trained models integrated into the MonoTransQuest architecture. Subsequently, we calculate the average of the cumulative score, divided by the number of pre-trained models, resulting in the ensembleTQ score. Finally, we compute the Spearman and Pearson correlation scores for the ensembleTQ score, providing a comprehensive evaluation of our ensemble approach.

% As shown in Table XXXXXX,  the ensembleTQ provides better performance for all the language pairs compared to the single model configuration. 

\section{Result and Discussion}

% The conducted experiments focus on language pairs categorized as mid-resourced and low-resourced datasets. 
The research is divided into two distinct settings, as outlined in Sections 5.1 and 5.2. The primary evaluation metric employed in this study is the Spearman correlation coefficient.

As shown in Table \ref{table_mymodels}, is notable that the baseline model does not surpass our proposed approaches in terms of Spearman correlation scores in most cases. This outcome underscores the specific strengths and limitations associated with different model architectures. Complementing the Spearman correlation analysis, the examination of Pearson correlation scores further enriches the assessment. The MonoTQ-InfoXLM-large model consistently exhibits superior Pearson correlation scores across a majority of the language pairs, accentuating its robust performance characteristics. 

From our experiment results, as shown in Table \ref{table_mymodels}, it's notable that the single-model configuration of MonoTQ-InfoXLM-large and MonoTQ-XLMV outperform ensemble-TQ for the majority of the language pairs. Observing the results outlined in both Table \ref{table_mymodels} and Table \ref{tab2_allmodels}, it becomes evident that MonoTQ-InfoXLM-large and MonoTQ-XLMV not only outperform other systems among our own proposed approaches, they also exhibit a competitive performance with the best-performing system in the WMT23 sentence-level shared-task. MonoTQ-InfoXLM-large shows a very close Spearman correlation score with the winning system of the WMT23 sentence-level task for the En-Gu, En-Hi and En-Ta language pairs. Also, MonoTQ-XLMV shows the highest Spearman correlation score for the En-Te language pair. This observation raises the question that do the practice of ensembling always guarantees performance enhancement. Table \ref{tab3_memory} presents the memory requirements of both ensemble approaches and single-model configurations. Interestingly, in most cases ensemble models demand significantly more memory space than single-model setups, despite only offering a marginal boost in performance. This observation prompts us to reconsider the efficiency of employing ensemble models.  

The conducted experiments across mid-resourced and low-resourced language pairs unravel intricate performance dynamics among various models.

\begin{table}
    \centering
    \begin{tabular}{|c|p{0.4\linewidth}|p{0.3\linewidth}|} \hline % Adjust the width as needed
         \textbf{No.} & \textbf{Name} & \textbf{DiskFootPrint (Bytes)}\\ \hline 
          & & \\
         1 & Unbabel-IST & 42,868,104,221\\
         & & \\
         2 & IOL Research & 2,357,242,105\\ 
          & & \\
         3 & HW-TSC & 27,730,527,504\\ 
          & & \\
         4 & MMT & 2,448,132,038\\
          & & \\
         5 & SurreyAI-ensembleTQ & 7,945,689,496\\ 
         6 & SurreyAI-MonoTQ-XLMV & 3,221,225,472\\ 
         7 & SurreyAI-MonoTQ-InfoXLM-large & 2,362,232,012\\ 
         8 & SurreyAI-MonoTQ-XLMR-large & 2,254,857,830\\ \hline
    \end{tabular}
    \caption{Rows 1-5 display the disk footprint of ensemble model submissions related to the sentence-level task for WMT23. Meanwhile, Rows 6-8 present the disk footprint of our TQ models with single model configuration.}
    \label{tab3_memory}
\end{table}

\section{Conclusion}
This paper comprehensively evaluates the proposed architecture within the context of sentence-level direct quality assessment, employing diverse encoder-based pre-trained models. Our investigation notably highlights the enhanced performance attributed to the MonoTQ-InfoXLM-large, which surpasses the other configuration approaches, namely MonoTQ-XLMV, ensembleTQ strategy and MonoTQ-XLMR-large. While our outcomes in the WMT23 sentence-level Direct Assessment task did not attain peak performance, they nevertheless exhibited a marked improvement over the baseline and showed notable performance scores close to the winning systems.

Looking ahead, our research trajectory anticipates a continued exploration of quality estimation employing large language models. This involves further experimentation encompassing a broader spectrum of low-resourced language pairs. These forthcoming endeavours aspire to deepen our insights into the intricacies of direct quality assessment and contribute to advancing the frontiers of natural language processing. Also, we are focused on continuing the experimentation of pre-trained language models incorporated into different QE frameworks.

\bibliography{CameraRead_SurreyAI_V3}
\bibliographystyle{acl_natbib}

\end{document}